\documentclass[lettersize,journal]{IEEEtran}
\usepackage{amsmath,amsfonts}
\usepackage{algorithmic}
\usepackage{algorithm}
\usepackage{pifont}
\usepackage{xcolor}
\usepackage{array}
\usepackage[caption=false,font=normalsize,labelfont=sf,textfont=sf]{subfig}
\usepackage{textcomp}
\usepackage{stfloats}
\usepackage{url}
\usepackage{verbatim}
\usepackage{graphicx}
\usepackage{cite}
\usepackage{booktabs}
\usepackage{multirow}
\usepackage{color}
\usepackage[backref]{hyperref}

\newcommand{\etal}{\textit{et al}.}

\hypersetup{
colorlinks=true,
linkcolor=black,
citecolor=black
}

\begin{document}

\title{Learning Attribute-aware Representations for Few-shot Scene Text Segmentation}

\author{Yifan~Tang,
        Chenming~Li,
        Chengxu~Liu,
        Yuanting~Fan,
        Dangfeng~Yang, 
        Yong~Huang, 
        Cun~Xin, 
        Yu~Li,
        Xingsong~Hou,
        and~Xueming~Qian

\thanks{This work was supported by the NSFC under Grant 62272380.}

\thanks{Chenming Li, Xingsong Hou and Yifan Tang are with the School of Information and Communication Engineering, Xi'an Jiaotong University, Xi'an 710049, China (e-mail: 3123352041@stu.xjtu.edu.cn; houxs@mail.xjtu.edu.cn; tangyifan@stu.xjtu.edu.cn).}
\thanks{Yuanting Fan is with the School of Software Engineering, Xi’an Jiaotong University, Xi’an 710049, China (e-mail: retofan@stu.xjtu.edu.cn).}
\thanks{Chengxu Liu is with the School of Software Engineering, Xi’an Jiaotong University, Xi’an 710049, China, and also with Shaanxi Yulan Jiuzhou Intelligent Optoelectronic Technology Company Ltd., Xi’an 710000, China (e-mail: chengxu.liu@xjtu.edu.cn).}
\thanks{Dangfeng Yang, Yong Huang, Cun Xin, and Yu Li are with Power China Northwest Engineering Corporation Limited, Xi’an 710100, China (e-mail: 33687571@qq.com; huangy@nwh.cn; xincun0904@163.com; liyu@nwh.cn).}
\thanks{Xueming Qian is with the Ministry of Education Key Laboratory for Intelligent Networks and Network Security, School of Information and Communication Engineering, Xi'an Jiaotong University, Xi'an 710049, China, and also with SMILES LAB, Xi'an Jiaotong University, Xi'an 710049, China (e-mail: qianxm@mail.xjtu.edu.cn).}
}



\maketitle

\begin{abstract}
Supervised scene text segmentation has achieved notable progress in recent years. However, its development is largely constrained by the scarcity of high-quality datasets and the high cost of pixel-level annotations. To address this limitation, we explore few-shot learning for text segmentation and propose TSAL, an attribute-aware few-shot framework that leverages a pre-trained CLIP model to learn transferable text attributes for segmentation. Our framework comprises two complementary branches: I) a Visual-Guided Branch that extracts semantic and textural features for foreground text and background regions, respectively, and II) an Adaptive Prompt-Guided Branch that employs learnable prompt templates to capture diverse text attributes with minimal data dependence. To effectively align textual attributes with visual representations, we further introduce an Adaptive Feature Alignment~(AFA) module, which aligns learnable attribute tokens with visual features and prompt prototypes, enabling the model to capture both general and distinctive textual characteristics. As a result, TSAL can accurately segment text regions using only a few annotated samples. Extensive experiments demonstrate that our method achieves state-of-the-art performance across several public text segmentation benchmarks under few-shot settings and exhibits strong generalization to text-related tasks.
\end{abstract}


\begin{figure}[t]
	\centering
   \includegraphics[width=1.0\linewidth]{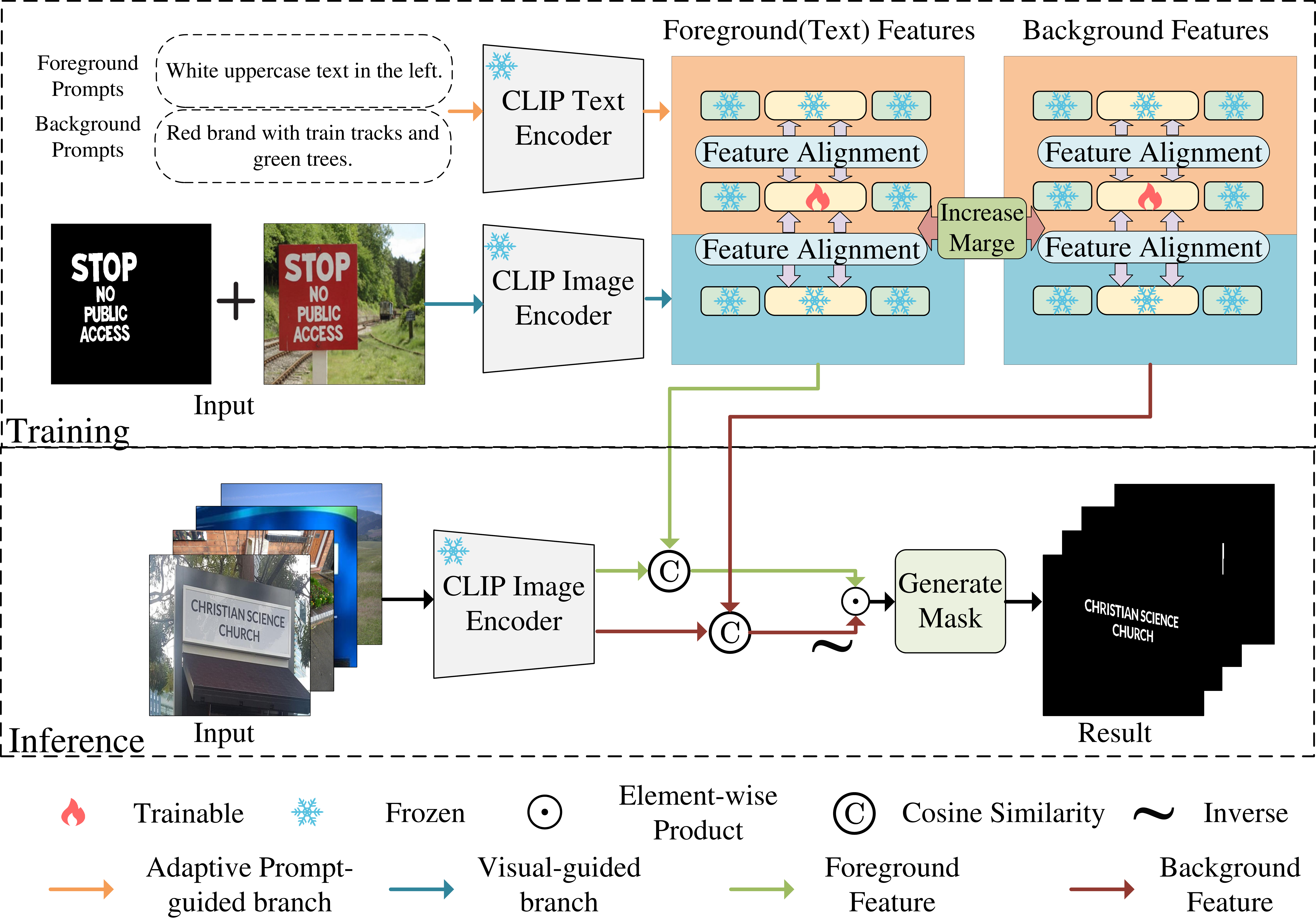}
        \caption{Training and Inference Pipeline of TSAL under 1-shot setting. During training, we optimize the foreground and background prompts to learn attribute features. During inference, we calculate the similarity between the optimized features and the input image, obtaining the final result.}
		\label{fig: teaser}
\end{figure}

\section{Introduction}
\IEEEPARstart{S}{cene} text segmentation aims to extract text from images captured in various scenarios. The extracted results are useful for applications such as text erasing~\cite{conrad2021two,zdenek2020erasing}, text recognition~\cite{atienza2021vision}, and text editing~\cite{li2024blip}. Advanced methods~\cite{yu2025eaformer, wang2023textformer} require large datasets for optimal performance. Nevertheless, pixel-level text annotation is expensive and time-consuming. Well-known datasets like COCO\_TS~\cite{bonechi2019coco_ts} and MLT\_S~\cite{bonechi2020weak} are large in size but lack precise annotations. Meanwhile, datasets with more accurate annotations, such as TextSeg~\cite{xu2021rethinking} and Total-Text~\cite{ch2017total}, are relatively small in scale. The above issues hinder the progress of scene text segmentation methods. 
Therefore, exploring effective methods to enhance text segmentation performance with limited samples is meaningful and valuable.

Generalized Few-shot Semantic Segmentation~(FSS) aims to identify pixels of novel classes using guidance from a limited number of labeled examples. Recent methods~\cite{tian2020prior, lang2022learning, leng2025multi, zhang2023rpmg} leverage prior information from pixel relationships between the support set and query image to guide the decoder, and design modules to aggregate contextual information across different scales. However, they still require large amounts of data and substantial training costs to recognize unseen categories. Additionally, these methods are designed for general category segmentation tasks and show limitations when applied to text segmentation, where texts are frequently obscured by complex background clutter and exhibit textures similar to those of the surrounding regions~\cite{ren2022looking}.

Vision-language models~\cite{radford2021learning, jia2021scaling, bao2021vlmo} advance rapidly and demonstrate strong potential for few-shot semantic segmentation~\cite{yin2024hierarchy}. Contrastive Language-Image Pretraining~(CLIP)\cite{radford2021learning} maps images and text into high-dimensional spaces using two structurally similar encoders, leveraging prior information for few-shot learning\cite{wang2024rethinking}. WinCLIP~\cite{jeong2023winclip} and PromptAD~\cite{li2024promptad} applies CLIP to anomaly detection, making prompts more adaptable and ensuring that learnable prompts capture visual features with the highest relevance. Filo~\cite{gu2024filo} incorporates additional prior knowledge and utilizes Large Language Models~(LLMs) to generate fine-grained prompts and make effective progress. These advancements demonstrate that optimizing prompts can enhance CLIP’s few-shot segmentation performance. Unfortunately, due to the highly complex topological structure of text and its deep mixture with backgrounds, existing methods do not fully leverage the advantages of CLIP, such as its rich attribute priors and efficient knowledge adaptation via soft prompts, both enabled by large-scale text-image alignment pretraining.

To address these challenges, we propose a few-shot scene text segmentation method named TSAL, which leverages CLIP’s text-visual matching capabilities to incorporate prior knowledge. 
As shown in Fig.~\ref{fig: teaser}, our goal is to optimize learnable prompts with only few images and directly use the optimized feature vectors for text segmentation.
Specifically, our model consists of a visual-guided branch and an adaptive prompt-guided branch. In the visual-guided branch, the image encoder extracts and stores foreground~(text) and background features from few-shot images. In the adaptive prompt-guided branch, we design effective prompt templates to learn the general and unique attributes of foreground and background. To further enhance the generalization of prompts, we propose Adaptive Feature Alignment module~(AFA). AFA leverages visual features to help adaptive prompts capture the texture information of real data, while using prompt prototypes to learn rich semantic information. During inference, we compute the similarity between the optimized tokens and the input images to obtain foreground and background score maps. Finally, they are fused to generate the mask.

The main contributions are summarized as follows:
\begin{itemize}
\item 
To the best of our knowledge, 
we are the first to explore scene text segmentation under few-shot settings and utilize CLIP's priors to improve performance.
\item Our adaptive prompt-guided branch and Adaptive Feature Alignment module enable prompts to capture the unique attributes of text, integrating with the visual branch to achieve precise text segmentation.

\item Experiments show that our simple and efficient method achieves impressive performance on multiple scene text segmentation datasets. 
\end{itemize}

The rest of the paper is organized as follows. In Sec.~\ref{sec:related}, we review the related works. In Sec.~\ref{sec:method}, we describe our TSAL in detail. In Sec.~\ref{sec:experiments}, we complete the comparison experiments and the ablation study. In Sec.~\ref{discussion}, we discuss the parameter settings and the model details.

\section{Related Work}
\label{sec:related}
\subsection{Scene Text Segmentation}
Scene text segmentation aims at predicting fine-grained masks for texts in scene images. Recent deep learning-based methods can be roughly divided into fully supervised learning methods~\cite{xu2021rethinking,xu2022bts,ren2022looking,wang2023textformer,yu2023scene,yu2025eaformer} and weakly supervised learning methods~\cite{wang2021semi, zu2023weakly}.

Fully supervised learning methods rely on large, high-quality labeled datasets to ensure model performance. Typically, TexRNet~\cite{xu2021rethinking} dynamically activates low-confidence regions by leveraging high-confidence features to improve text segmentation integrity. PGTSNet~\cite{xu2022bts} integrates a text detection module for text localization. ARM-Net~\cite{ren2022looking} combines global and local features while minimizing background noise. TextFormer~\cite{wang2023textformer} employs a multi-level Transformer to capture textual features. TFT~\cite{yu2023scene} uses a text-focused module to guide the model to locate text regions, while EAFormer~\cite{yu2025eaformer} enhances segmentation by emphasizing text edges. Though effective in extracting texture features, these methods are limited by dataset quality. To address the issue of limited data, WASNet~\cite{xiedataset} introduces a novel synthetic dataset WAS-S. However, the quality of the synthetic dataset still differs from that of real data.

Due to the lack of massive pixel-level labeled data for supervised training, Wang~\etal~\cite{wang2021semi} proposes semi-supervised learning methods using low-cost polygon masks. TAR~\cite{zu2023weakly} uses a text detector to locate text areas and a text recognizer to generate pseudo labels. Although these methods reduce the reliance on fine-grained annotations, they still require a large amount of data for training. Hi-SAM~\cite{ye2024hi} extends the Segment Anything Model to text segmentation by establishing hierarchical feature representation for document analysis. Our method leverages CLIP prior knowledge, enabling precise text segmentation with few images.

\subsection{Few-Shot Semantic Segmentation}
Few-Shot Semantic Segmentation~(FSS) aims at pixel-level segmentation of new classes with less labeled data~\cite{fateh2024msdnet}. Typically, NTRENet~\cite{liu2022learning} refines query features by progressively filtering out background information. 
BAM~\cite{lang2022learning} enhances segmentation accuracy by combining a base learner for known categories with adaptive integration for target regions. Most recent methods use CLIP's prior knowledge to assist in segmentation.
PFENet~\cite{tian2020prior} introduces Prior Generation, which combines support features with prior information to refine query features. 
Wang~\etal~\cite{wang2024rethinking} utilizes the semantic alignment capability of CLIP to locate the target category. 
WinCLIP~\cite{jeong2023winclip} uses CLIP's image and text encoders to locate anomalous regions through text prompt learning. Their success further demonstrates the importance of incorporating prior information in effectively enhancing model performance.

\begin{figure*}[t]
	\centering
   \includegraphics[width=1.0\linewidth]{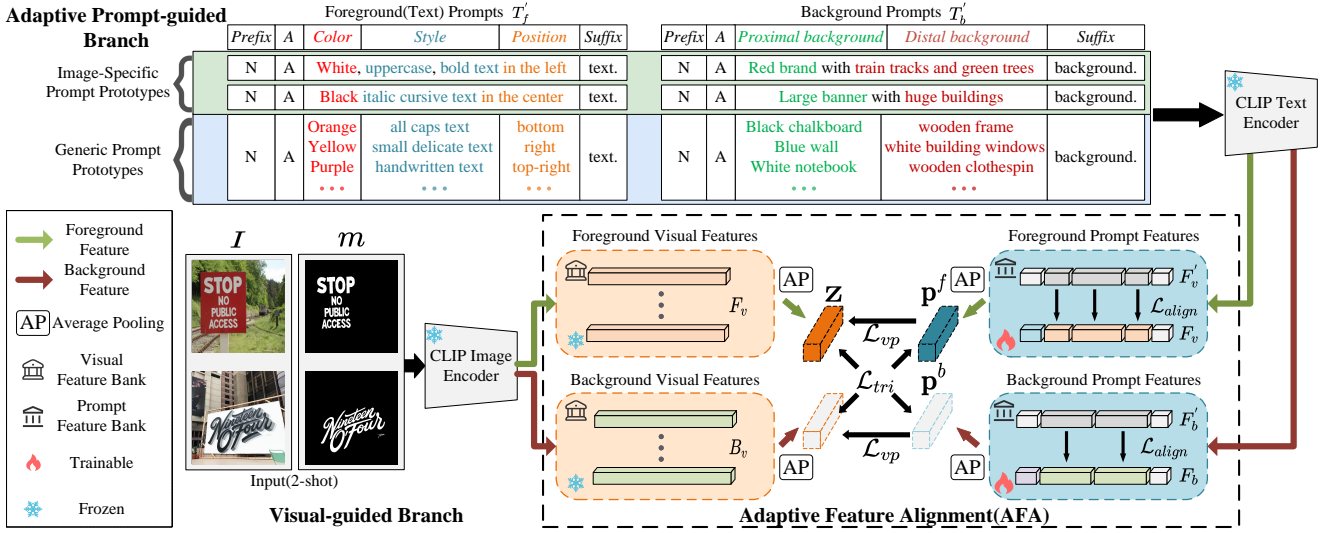}
        \caption{Overview of TSAL~(2-shot): the Visual-guided branch extracts and stores foreground and background features in visual feature bank for inference. The adaptive prompt-guided branch design effective templates for adaptive prompts and utilizes LLMs to generate fine-grained descriptions for each input image. AFA aligns the adaptive prompts with visual features and prompt prototypes, enabling them to capture rich semantic information. The optimized prompt features are stored in the prompt feature bank for inference.}
		\label{fig: overview}
\end{figure*}
\subsection{Vision-Language Model}

Thanks to the strong cross-modal understanding of vision-language models~(VLMs), many methods~\cite{kim2021vilt, li2019visualbert, 11207741} have been applied to downstream tasks. Recent advances are driven by contrastive learning~\cite{chen2020simple} and vision transformers~\cite{alexey2020image}. CLIP~\cite{radford2021learning}, pre-trained on large-scale image-text pairs, demonstrates impressive zero-shot capabilities. OpenCLIP~\cite{cherti2023openclip} further enables training on datasets like LAION-400M/5B~\cite{schuhmann2021laion, schuhmann2022laion}. With strong semantic alignment, CLIP maps image-text features into high-dimensional spaces. Combined with prompt learning, it significantly boosts performance on downstream tasks~\cite{luddecke2022image, li2023clip}. LSeg~\cite{li2022language} encodes labels via text embeddings for dense predictions. CoOp~\cite{zhou2022learning} and CoCoOp~\cite{zhou2022conditional} optimize or generate prompts for better domain generalization, highlighting the importance of prompt tuning for CLIP adaptability.

\section{METHODOLOGY}
\label{sec:method}
In this section, we present the proposed method. After introducing the preliminaries and problem formulation, we provide an overview of the framework, which consists of a visual-guided branch, an adaptive prompt-guided branch, and an adaptive feature alignment module. The inference procedure is described at the end.
\subsection{Preliminaries and Problem Formulation}
CLIP~\cite{radford2021learning} learns the relationship between images and text using contrastive learning, mapping visual and linguistic concepts in a shared embedding space.
Specifically, giving an unknown image $I$, and $K$ prompts ${p_1,p_2,...,p_K}$, CLIP can predict the distribution of $I$ belonging to $K$ prompts:
\begin{equation}
p(y|I)=\frac{\exp<f(I),g(p_y)/\tau>}{\sum_{i=1}^K\exp<f(I),g(p_i)/\tau>},
\end{equation}
where $<\cdot>$ refers to the cosine similarity. $f(\cdot)$ and $g(\cdot)$  denote the image encoder and text encoder respectively. $\tau$ is the temperature hyper-parameter.
This allows CLIP to rank the given prompts based on their relevance to the image, effectively predicting which prompt best describes the content of the image and providing a more accurate alignment between textual and visual information.

Inspired by the success of prompt learning in natural language processing~(NLP)~\cite{shin2020autoprompt,jiang2020can}, CoOp~\cite{zhou2022learning} introduces the prompt learning paradigm to few-shot classification, aiming to automatically learn efficient prompts for CLIP.
Specifically. CoOp embeds learnable tokens in the prompt:
\begin{equation}
{p}_i=[\mathbf{L}_1][\mathbf{L}_2]\ldots[\mathbf{L}_{N}][class_i],
\end{equation}
where $[\mathbf{L}_1][\mathbf{L}_2]\ldots[\mathbf{L}_{N}]$ are learnable tokens, $[class_i]$ is $i$\text{-}th class name which is not trainable. Prompt learning successfully transferring CLIP's extensive prior knowledge to improve performance on these tasks.

Since scene text segmentation is a binary classification task, our method differs from general Few-shot Semantic Segmentation.
Given a dataset $D$, it is divided into a support set $D_s$ and a query set $D_q$. $D_s$ contains few-shot image-mask pairs. During training, our model only utilizes $D_s$ to enhance the ability to distinguish between foreground and background. During inference, we predict the images in $D_q$ and obtain final masks $R$. 
We build our method upon CLIP~\cite{radford2021learning}, which aligns images and text in a shared embedding space using contrastive learning. Prompt learning techniques such as CoOp~\cite{zhou2022learning} enable CLIP to adapt to specific tasks via learnable textual representations. In our case, prompts are designed to guide segmentation by bridging visual features and text semantics. Detailed formulations are provided in the supplementary material.
\subsection{Overview}
An overview of our method is illustrated in Fig.~\ref{fig: overview}. 
Our model consists of two branches: the visual-guided branch, and the adaptive prompt-guided branch. For the vision-guided branch, the input image $I$ is fed into the CLIP Image Encoder, and its mask $m$ is used to separate the features into foreground features and background features, which are then stored in the visual feature bank. For the adaptive prompt-guided branch, we first utilize LLMs to generate fine-grained textual attribute descriptions and background attribute descriptions for $I$ based on the designed template. Additionally, to ensure the prompts adapt to various real-world scenarios, we generate a series of generic descriptions for each attribute. These descriptions are then fed into the CLIP Text Encoder for feature extraction to obtain the representations of foreground and background prototypes. To ensure that the prompts accurately represent real-world data, we introduce Adaptive Feature Alignment module~(AFA). By aligning learnable tokens with visual feature and prompt prototypes, AFA enables adaptive prompts to capture valuable information. We store the prompt prototypes and adaptive prompt features in the prompt feature bank. During inference, we compute the cosine similarity between the stored features in the visual feature bank and the prompt feature bank with the input image to obtain the score map and generate the mask.


\subsection{Visual-guided Branch}
\label{Visual-guided Branch}
To make full use of background information to enhance text localization, we utilize the mask $m$ of the input image $I$ to separately extract foreground features ${F}^{v} $ and background features ${B}^{v}$:
\begin{align}
{F}^{v}= f(I \odot m ), \quad
{B}^{v}= f(I \odot \overline{m}),
\end{align}
where $\odot$ is the element-wise product operation. $\overline{m}$ denotes the inverse of GT mask. $f(\cdot)$ is the CLIP Image Encoder. Then they are stored in the visual feature bank separately. 

During inference, the query image is encoded by the same image encoder as $Q$. We compare $Q$ with the visual feature bank to obtain foreground and background scores map respectively:
\begin{align}
{V}^{f}_{ij}=\max_{{q}\in{Q}}\frac{1}{2}(<F^v_{ij},{q}>), \\
{V}^{b}_{ij}=\max_{{q}\in{Q}}\frac{1}{2}(<B^v_{ij},{q}>).
\end{align}
where $ij$ is the feature index in the visual feature bank, and $<\cdot>$ denotes the cosine similarity. In practice, we use two layers of intermediate features as memories to obtain two score maps for each query image and then average the two score maps to obtain the final score map ${V}^{f}$ and ${V}^{b}$.

\subsection{Adaptive Prompt-guided Branch}
\label{AP}
The complex background and diverse text styles present many challenges for scene text segmentation.
Relying solely on the visual-guided branch provides only coarse localization, which leads to inaccurate detection of foreground regions. Therefore, it is essential to use the CLIP Text Encoder and design appropriate prompts to assist in detecting target~\cite{li2024promptad, gu2024filo}. 

For some challenging artistic fonts, the text exhibits strong stylistic features~\cite{xiedataset}. Additionally, the position and color of the text vary widely across different images. To enable the prompt to learn the above attributes of the text, we designed the following template:
\[
\setlength{\tabcolsep}{4pt}
\begin{array}{c@{}c@{}c@{}c@{}c@{}c@{}c@{}c@{}c}
T_{f}& = &[A] & [color] & [style] &text & [position],  \\
    & & \textcolor{blue}\downarrow\! & \textcolor{blue}\downarrow\! & \textcolor{blue}{\downarrow}\! & \textcolor{blue}\downarrow\! &\textcolor{blue}\downarrow\! \\
    T_f^{'} &=& \text{A} & \{\text{Black}\} &  \{\text{italic cursive}\} &\text{text} &\{\text{in the center}\}, \\
\end{array}
\]
where $T_{f}$ is the learnable prompt and $T_f^{'}$ is a prompt prototype. $[color]$, $[style]$, and $[position]$ respectively refer to different attributes of the text and they are replaced with learnable tokens. as shown in Fig.~\ref{fig: overview}. Note that the ``\textit{prefix}'' and ``\textit{suffix}'' in the figure are fixed fields used only for feature alignment during training and have no actual meaning. $[A]$ is an adaptive learnable prefix designed to capture implicitly defined text attributes and dynamically adapt to information within visual features. The setting of $[A]$ and learnable tokens can be found in ``Hyper-parameter Analysis" section.


For the background, we analyze all the images in the datasets and find that defining simple background prompt templates is not enough. We use the following template:
\[
\setlength{\tabcolsep}{3pt}
\begin{array}{c@{}c@{}c@{}c@{}c@{}c@{}c@{}c}
T_{b} &=& [A] & [proximal] & with &[distal], &  \\
    & & \textcolor{blue}\downarrow\! & \textcolor{blue}\downarrow\! & \textcolor{blue}{\downarrow}\! & \textcolor{blue}\downarrow\!  \\
    T_b^{'} &=& \text{A} & \{\text{Large banner}\} & \text{with} & \{\text{huge buildings}\}, \\
\end{array}
\]
where $[proximal]$ refers to proximal background and $[distal]$ refers to distal background. They are replaces with learnable tokens. If we use only  $[proximal]$ or $[distal]$, the localization of the background region becomes inaccurate, ultimately affecting the final text segmentation. Experiments show the effectiveness of the template.

To enable the prompts to describe foreground and background information in a more fine-grained manner, we use LLMs to generate image-specific prompts for the support images. However, using only prompts adapted to support images cannot fully cover the diverse real-world scenarios in the query set. Therefore, we define multiple candidate words for each attribute and generate generic prompts by combining them. These prompts are fed into the CLIP Text Encoder to obtain foreground and background features:
\begin{equation}
    \begin{aligned}
        F_f &= g(T_f),   & F_f^{'} &= g(T_f^{'}), \\
        F_b &= g(T_b),   & F_b^{'} &= g(T_b^{'}),
    \end{aligned}
\end{equation}
where ${g}(\cdot)$ is the CLIP Text Encoder.
$F_f$ and $F_f^{'}$ are the foreground adaptive prompt feature and prototype feature, respectively, while $F_b$ and $F_b^{'}$ are the background adaptive prompt feature and prototype feature.
 \begin{table*}[t]
    \centering
    \caption{Comparison of different methods in FgIoU/AUROC on TextSeg, Total-Text datasets and ICDAR13 FST. For methods that only output binary masks, AUROC cannot be computed and is therefore omitted. SegGPT is limited to the 1-shot setting and thus is not included in the multi-shot evaluation. The best and second-best results are respectively marked in red and blue.}
     \label{tab: Primary}
     \vspace{-2mm}
    \begin{tabular}{lccc|ccc|ccc}
        \toprule
        \multirow{2}{*}{Methods}  & \multicolumn{3}{c|}{TextSeg} & \multicolumn{3}{c}{Total-Text} & \multicolumn{3}{c}{ICDAR13 FST} \\
          & 1-shot & 2-shot & 4-shot & 1-shot & 2-shot & 4-shot & 1-shot & 2-shot & 4-shot \\
        \midrule
         BAM~\cite{lang2022learning} &10.95/-& 11.31/-& 12.89/- & 10.35/- & 10.86/- & 11.33/- & 8.71/-&8.86/- & 9.80/- \\
        HDMNet~\cite{peng2023hierarchical}&11.84/- & 12.11/- & 12.20/- & 10.17/- & 11.19/- & 12.23/-&9.02/- & 9.23/- & 9.80/-  \\
        SegGPT~\cite{wang2023seggpt}& 30.63/- & -/- & -/- & 21.13/- &-/- & -/- &  36.37/- & -/- & -/- \\
        TexRNet~\cite{xu2021rethinking} & 16.93/78.34 & 17.74/78.59 & 18.62/78.98 & 10.14/73.16 & 10.45/73.44& 10.52/73.63 &  14.17/73.20 & 14.20/73.58 & 14.54/74.11\\
        TFT~\cite{yu2023scene}& 17.74/79.44 & 19.85/79.56 & 20.72/79.83 &10.63/73.25 &11.17/73.64 &11.84/74.03& 14.45/74.02 & 14.79/74.79& 14.88/74.80 \\
        Zegclip~\cite{zhou2023zegclip}& 34.12/81.14 & 35.56/81.45 & 36.78/81.59 & 29.45/81.14&30.12/81.65&30.45/81.94 &38.15/82.81&38.59/83.14 & 38.74/83.45\\
         CAT-Seg~\cite{cho2024cat}& 37.25/83.85 & 38.40/84.20 & 39.55/84.60 & 32.60/83.80&33.35/84.15&34.10/84.50 &40.25/85.05&40.80/85.40 & 41.35/85.75\\
        WinCLIP~\cite{jeong2023winclip} &52.82/87.47 & 53.43/\textcolor{blue}{89.35} & 53.96/90.01 & 45.11/88.78 & 47.12/89.32 & 47.21/89.77 & 43.43/86.75 & 43.79/86.87 & 44.45/87.44\\
        PromptAD~\cite{li2024promptad}&\textcolor{blue}{53.31}/\textcolor{blue}{88.60} & \textcolor{blue}{54.72}/89.22 & \textcolor{blue}{55.74}/\textcolor{blue}{90.56} & \textcolor{blue}{46.31}/\textcolor{blue}{88.91} & \textcolor{blue}{47.72}/\textcolor{blue}{89.85} &\textcolor{blue}{47.84}/\textcolor{blue}{90.03} &\textcolor{blue}{45.98}/\textcolor{blue}{87.44} & \textcolor{blue}{46.65}/\textcolor{blue}{88.46} &\textcolor{blue}{46.90}/\textcolor{blue}{88.74}\\
        \midrule
            TSAL& \textcolor{red}{66.63}/\textcolor{red}{95.17}&  \textcolor{red}{67.73}/\textcolor{red}{95.65} & \textcolor{red}{68.24}/\textcolor{red}{96.21} & \textcolor{red}{58.13}/\textcolor{red}{93.66}&\textcolor{red}{58.56}/\textcolor{red}{94.54}  & \textcolor{red}{59.74}/\textcolor{red}{94.72} & \textcolor{red}{62.63}/\textcolor{red}{94.68} & \textcolor{red}{63.33}/\textcolor{red}{94.75} &  \textcolor{red}{63.65}/\textcolor{red}{95.12} \\
        \bottomrule
    \end{tabular}
    \vspace{-2mm}
\end{table*}

 \begin{table}[t]
 \centering
     \caption{Comparison of different methods in FgIoU/AUROC on ICDAR13 FST and BTS datasets. The best and second-best results are
respectively marked in red and blue.}
     \label{tab:secondary}
     \vspace{-2mm}
    \begin{tabular}{lccc}
        \toprule
        \multirow{2}{*}{Methods}& \multicolumn{3}{c}{BTS} \\
           & 1-shot & 2-shot & 4-shot  \\
        \midrule
        BAM~\cite{lang2022learning}  & 9.12/- & 9.34/- & 9.95/-\\
        HDMNet~\cite{peng2023hierarchical}& 9.33/- & 9.43/- & 9.62/- \\
        SegGPT~\cite{wang2023seggpt}& 29.46/- &-/- & -/- \\
        TexRNet~\cite{xu2021rethinking} & 15.47/73.34 & 16.54/74.59& 17.65/74.87 \\
        TFT~\cite{yu2023scene} & 16.54/74.25 & 18.75/75.64 & 18.98/76.14 \\
        Zegclip~\cite{zhou2023zegclip} & 31.20/82.45 & 32.10/82.85 & 33.45/83.20 \\
        CAT-Seg~\cite{cho2024cat} & 33.85/83.90 & 34.80/84.30 & 35.95/84.70 \\
        WinCLIP~\cite{jeong2023winclip}& 39.45/87.45 & 40.21/88.18 & 41.54/89.30\\
        PromptAD~\cite{li2024promptad} & \textcolor{blue}{41.78}/\textcolor{blue}{88.12} & \textcolor{blue}{42.56}/\textcolor{blue}{89.56} & \textcolor{blue}{43.74}/\textcolor{blue}{90.10} \\
        \midrule
        TSAL & \textcolor{red}{57.24}/\textcolor{red}{93.85}&\textcolor{red}{57.89}/\textcolor{red}{94.55}  & \textcolor{red}{58.32}/\textcolor{red}{94.64} \\
        \bottomrule
    \end{tabular}
     \vspace{-5mm}
\end{table}


\subsection{Adaptive Feature Alignment}
To enhance the generalization of adaptive prompts and enable $F_f$ and $F_b$ to learn the attribute features of the foreground and background, respectively, we propose the Adaptive Feature Alignment module~(AFA). 

To ensure that the adaptive prompts can learn diverse attribute features, we define the loss $\mathcal{L}_{align}$: 
\begin{equation}
\mathcal{L}_{align} = d(g(L), g(L^{'})),
\end{equation}
where $d(\cdot)$ represents euclidean distance. $L$ and $L^{'}$ represent the adaptive prompt feature and prototype feature, respectively. By utilizing $\mathcal{L}_{align}$, we constrain the learnable prompts to align with the prototype prompts. Therefore, the foreground prompts are able to capture rich semantic and stylistic information from the text, while the background prompts focus on learning complex contextual features.

Thanks to the support image features stored in the visual feature bank, aligning the adaptive features with the bank features allows the prompts to learn the corresponding visual distribution, thereby enhancing the robustness and flexibility of the adaptive prompts. Take the foreground as an example, we define the loss function as follows:
\begin{equation}\small
\mathcal{L}_{vp}=\mathbb{E}_{\mathbf{z}}\left[-log\frac{\exp(<\mathbf{z},{\mathbf{p}}^f/\tau>)}{\exp<\mathbf{z},{\mathbf{p}}^f/\tau>+\exp<\mathbf{z},\mathbf{p}^{b}/\tau>}\right],    
\end{equation}
where $\mathbf{z}$ is foreground visual feature $F^{v}$ after average pooling. $\mathbf{p}^f =\frac{\sum{g(T_f^n)}}N  $ is the average foreground prompt features, $\mathbf{p}^b =\frac{\sum{g(T_b^m)}}M  $ is the average background prompt features. $N$ and $M$ represent the number of foreground and background prompts, respectively. $\tau$ is the temperature hyper-parameter.


To enable the foreground and background prompts better to learn different semantic information separately, we minimize the distance between foreground visual features and prompt features while maximizing their separation from background prompt features. We apply the same loss function to both foreground and background prompts:
\begin{equation}
\mathcal{L}_{tri}=\mathbb{E}_{\mathbf{z}}[\max\left(\|\mathbf{z}-\mathbf{p}^{f}\|^{2}-\|\mathbf{z}-\mathbf{p}^{b}\|^{2} + \beta, 0\right)],
\end{equation}
where $\beta$ is set to $0.3$ following common practice in metric learning.
By utilizing $\mathcal{L}_{tri}$, the adaptive prefixes~($[A]$) learn attributes that are not explicitly defined by the prototype. 
This strategy allows the adaptive prompts to more effectively utilize CLIP's visual priors while also achieving greater adaptability to the text segmentation task. The optimized $F_f$, $F_f^{'}$, $F_b$ and $F_b^{'}$ are stored in the prompt feature bank for inference.

\begin{figure}[t]
	\centering
   \includegraphics[width=1.0\linewidth]{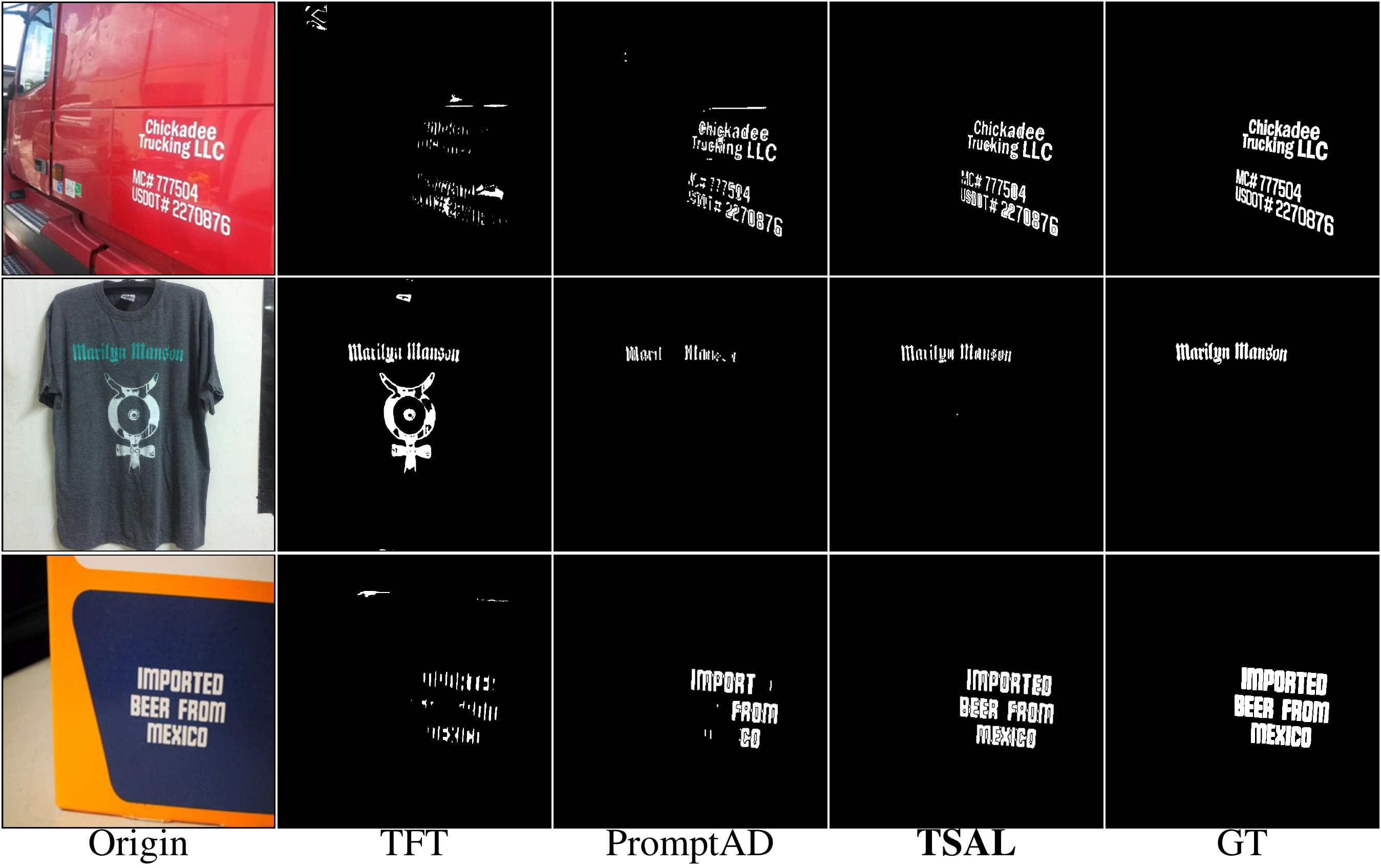}
   \vspace{-4mm}
        \caption{Visualization results under the 1-shot setting on the TextSeg, Total-Text, and IDAR13 FST datasets. To ensure fairness, PromptAD adopts our mask generator.}
		\label{fig: case}
\end{figure}

Our joint loss is defined as:
\begin{equation}
    L_{sum}=L_{align}+\alpha L_{vp}+L_{tri},
\end{equation}
where $\alpha$ is empirically set to 0.01, and both $L_{vp}$ and $L_{tri}$ explicitly consider foreground and background.

\subsection{Inference}
\label{Inference}
As noted in FG-CLIP~\cite{xie2025fg}, CLIP’s text encoder has a fixed token limit~(77 tokens), making it challenging to effectively process lengthy, fine-grained descriptions~(e.g.,~position/style/color). Sentence-level prompts condense semantic information more efficiently, avoiding truncation and improving generalization.
Similar to the visual-guided branch, we obtain the score map for the adaptive prompt-guided branch:
\begin{align}
P_f =softmax(\max(<P^{f}_i, Q>)), \\ 
P_b = softmax(\max(<{P}^{b}_i,Q>)),
\end{align}
where $P_f$ is the foreground score map. $P_b$ is the background score map. $P^{f}$ is the union of foreground prompt features. ${P}^{b}$ is the union of background prompt features. 
To refine the foreground region and remove noise from the background, we optimize the score map as follows:
\begin{align}
V^{'}_{f} = V^f\odot(1-V^b), \\
P^{'}_{f} = P_f\odot(1-P_b).
\end{align}
Then, we fuse them to obtain the final score $S$ using harmonic mean, which is more sensitive to smaller values~\cite{jeong2023winclip}:
\begin{equation}
S = \frac{V^{'}_{f} \cdot P^{'}_{f}}{V^{'}_{f} +P^{'}_{f}}.
\end{equation}
Finally, we use the mask generator to produce the result $R$, which is defined as:
\begin{equation}
  R = \text{FloodFill}(\psi_{\text{erode}}(\psi_{\text{dilate}}({\text{Canny(\textit{I})}}))  \cap B) \cap B,  
\end{equation}
where $B = (S > 0.5)$. $\text{Canny}(\mathit{I})$ denotes the Canny edge detection applied on the input image $\mathit{I}$. $\psi_{\text{dilate}}(\cdot)$ and $\psi_{\text{erode}}(\cdot)$ represent standard morphological dilation and erosion operations respectively. The $\text{FloodFill}(\cdot)$ operation fills closed regions delineated by the processed edges, while constraining the filling process to the foreground region $B$, thereby ensuring that only valid regions are retained.

\section{Experiments}
\label{sec:experiments}
In this section, we perform extensive experiments on four public datasets under diverse few-shot settings. We validate the effectiveness of our method through comprehensive comparisons with state-of-the-art~(SOTA) methods, qualitative visualizations, and ablation studies on key model components.
\subsection{Datasets and Metrics}
To demonstrate the validity of our method, we conduct extensive experiments using four publicly available text segmentation datasets~(TextSeg, Total-Text, ICDAR13 FST and BTS) under 1, 2, and 4-shot settings. 
TextSeg~\cite{xu2021rethinking} is a large-scale fine-annotated and multi-purpose text detection and segmentation dataset, collecting scene and design text with six types of annotations: word- and character-wise bounding polygons, masks, and transcriptions. It contains 4,024 images and involves two types of texts in these images: scene texts and design texts. Total-Text~\cite{ch2017total} is a text detection dataset that consists of 1,555 images with a variety of text types including horizontal, multi-oriented, and curved text instances. The training split and testing split have 1,255 images and 300 images, respectively. ICDAR13 FST~\cite{karatzas2013icdar} consists of 229 training images and 233 testing images, with word-level annotations provided.  BTS~\cite{xu2022bts} is a large-scale bi-lingual text segmentation dataset, which contains 14, 250 images with 44, 280 textlines and 209, 090 characters.
For fairness, we follow existing works~\cite{li2024promptad, wang2023textformer, yu2025eaformer} and use FgIoU and the Area Under the Receiver Operating Characteristic~(AUROC) as our evaluation metrics to validate the effectiveness of our method.

\begin{table}[t]
\centering
 \caption{Comparison with SOTA methods under many-shot settings in FgIoU and AUROC.}
 \label{tab:many-shot}
 \vspace{-2mm}
\begin{tabular}{lccc}
\hline
{Methods} & {Setting} & {FgIoU}$\uparrow$ \\ \hline
\textbf{TSAL}  & 1-shot & 66.43 \\ 
\textbf{TSAL}  & 4-shot &  \textcolor{red}{68.24}   \\ \hline
WinCLIP~\cite{jeong2023winclip} & 64-shot & 59.45 \\ 
PromptAD~\cite{li2024promptad}   & 64-shot &  62.46  \\ \hline
TexRNet~\cite{xu2021rethinking} &128-shot &  66.72 \\
TFT~\cite{yu2023scene} & 128-shot &67.12  \\\hline
\end{tabular}
\end{table}

\begin{figure}[t]
	\centering
\includegraphics[width=1.0\linewidth]{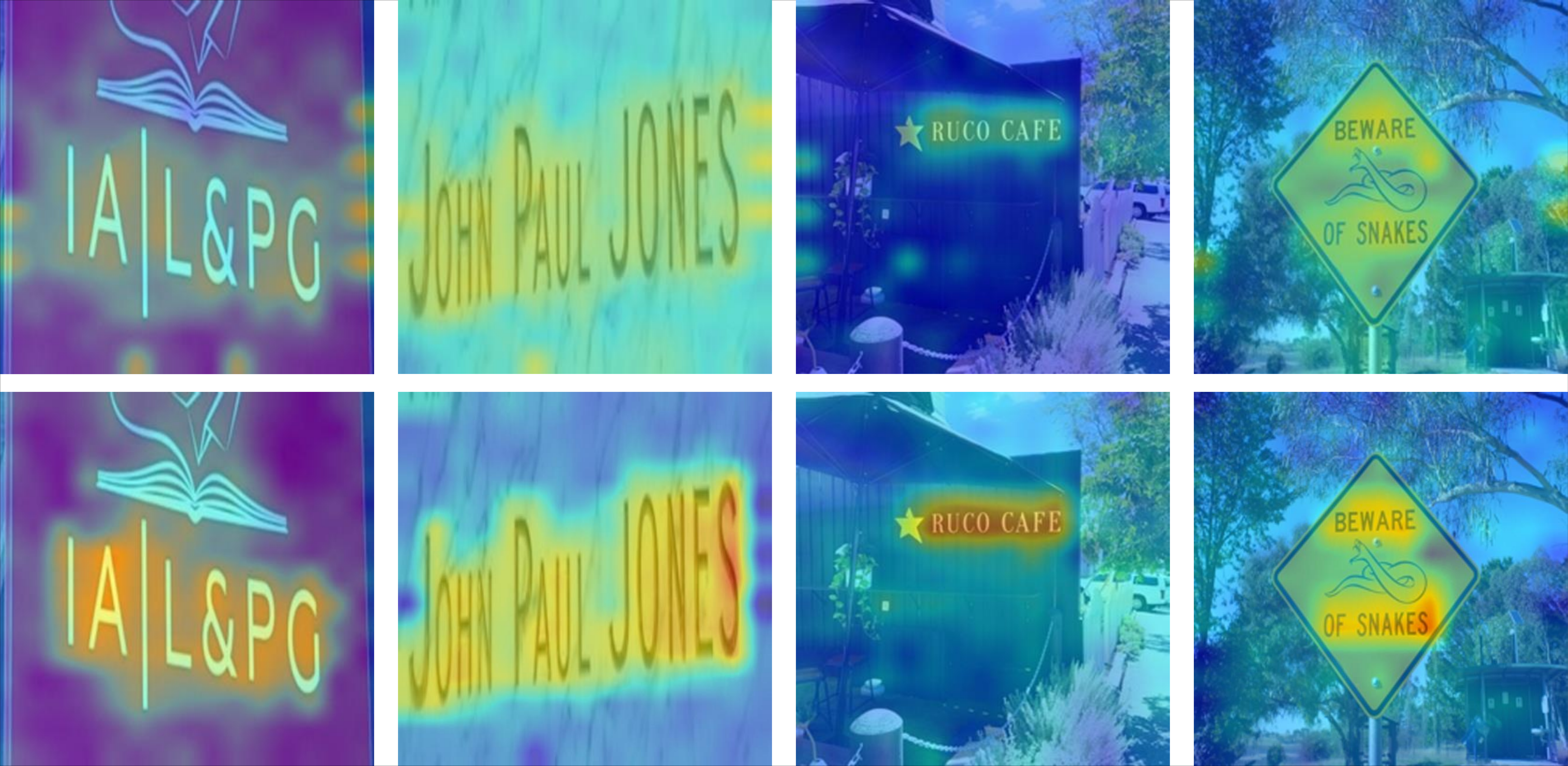}
        \caption{Visualization examples of AFA for different scenarios. The first row is \textit{w/o} AFA, the second row is \textit{w/} AFA.}
		\label{fig: AFA}
\end{figure}

\begin{table}[t]
\centering
\caption{Ablation study results for different modules.}
\label{tab:abla}
\vspace{-2mm}
\begin{tabular}{ccccc}
\hline

{Visual} &  {Prompt}  & {AFA} &{FgIoU$\uparrow$}&{AUROC$\uparrow$} \\ \hline

\textcolor{green}{\ding{51}}&  \textcolor{red}{\ding{55}}
  & \textcolor{red}{\ding{55}} &  62.45  &92.49   \\
 \textcolor{green}{\ding{51}}& \textcolor{green}{\ding{51}}&  \textcolor{red}{\ding{55}}
   &  63.75  &93.19 \\
  \textcolor{green}{\ding{51}} &   \textcolor{green}{\ding{51}}   & \textcolor{green}{\ding{51}}   & \textbf{66.63} &\textbf{95.17} \\
\hline
\end{tabular}
\end{table}

\subsection{Implementation Details}
We utilize the OpenCLIP~\cite{cherti2023openclip} implementation of CLIP, including its pre-trained parameters and the default hyper-parameter value for $\tau$. Following WinCLIP~\cite{jeong2023winclip}, we employed a CLIP model based on LAION-400M~\cite{schuhmann2021laion} with the ViT-B/16+~\cite{alexey2020image} architecture. We train our method for 20 epochs on an Nvidia GeForce RTX 3090 GPU. 
In each training session, we set a random seed for support image selection and prompt candidates. To ensure reliability, we average results across different seeds. Results from experiments with different random seeds are provided in the supplementary material.

\subsection{Comparison with state-of-the-art Methods}
To demonstrate the validity of TSAL, we compare the SOTA methods in the fields of general few-shot segmentation~(BAM~\cite{lang2022learning}, HDMNet~\cite{peng2023hierarchical}), few-shot object detection~(SegGPT~\cite{wang2023seggpt}, WinCLIP~\cite{jeong2023winclip}, PromptAD~\cite{li2024promptad}), text segmentation~(TexRNet~\cite{xu2021rethinking}, TFT~\cite{yu2023scene}), and related open-vocabulary CLIP adaptations (ZegClip~\cite{zhou2023zegclip}, CAT-Seg~\cite{cho2024cat}). Due to the lack of publicly available implementations, some SOTA methods are not included. All methods are reproduced in the released code.

\noindent\textbf{Quantitative comparison.} We evaluate the proposed method against state-of-the-art methods. The results are shown in Tab.~\ref{tab: Primary} and Tab.~\ref{tab:secondary}. 
HDMNet~\cite{lang2022learning} focuses on segmenting regular categories, but its performance is limited when segmenting text. TFT~\cite{yu2023scene} requires a large amount of labeled data for training and has limited performance with few input images. Compared to PromptAD, our method achieves 
 13.32\%, 13.01\%, and 12.5\% improvement on FgIoU under the 1, 2, and 4-shot settings. Our method designs unique learnable prompts for both the foreground and background separately, fully leveraging the mutual enhancement of foreground and background features, resulting in better performance. 
To demonstrate the generalization of TSAL, we conduct extensive experiments on the Total-Text~\cite{xu2021rethinking}, ICDAR13 FST~\cite{karatzas2013icdar} and BTS~\cite{xu2022bts}. Results show that TSAL performs impressively in a few-shot setting.

To demonstrate the effectiveness of our prompt learning strategy, we compare our 1-/4-shot results with other many-shot methods, as shown in Tab.~\ref{tab:many-shot}. Our 4-shot results perform better than other many-shot methods, even when compared to methods under 128-shot setting. Our method designs unique learnable prompts for both the foreground and background separately, fully leveraging the mutual enhancement of foreground and background features, resulting in better performance. More comparisons with SOTA methods are provided in the supplementary material.

\begin{table}[t]
\centering
\caption{Ablation study results on foreground prompt.}
\label{tab:fa}
\begin{tabular}{ccccc}
\hline

{Color} &  {Style}  & {Pos} & {Fine-grained}&{FgIoU$\uparrow$}\\ \hline
\textcolor{red}{\ding{55}}&  \textcolor{red}{\ding{55}}
  & \textcolor{red}{\ding{55}} &\textcolor{red}{\ding{55}}&  62.68  \\
\textcolor{green}{\ding{51}}&  \textcolor{red}{\ding{55}}
  & \textcolor{red}{\ding{55}} &  \textcolor{red}{\ding{55}}& 64.89    \\
 \textcolor{green}{\ding{51}}& \textcolor{green}{\ding{51}}&  \textcolor{red}{\ding{55}}&\textcolor{red}{\ding{55}}
   &   65.32  \\
  \textcolor{green}{\ding{51}} &   \textcolor{green}{\ding{51}}   & \textcolor{green}{\ding{51}}   & \textcolor{red}{\ding{55}}& 66.45  \\
   \textcolor{green}{\ding{51}} &   \textcolor{green}{\ding{51}}   & \textcolor{green}{\ding{51}}   & \textcolor{green}{\ding{51}}&\textbf{66.63}  \\
  
\hline
\end{tabular}
\end{table}

\begin{table}[t]
\centering
\caption{Ablation study results for background prompt.}
\label{tab:ba}
\begin{tabular}{cccc}
\hline
Distal&Proximal &Fine-grained &{FgIoU$\uparrow$} \\ \hline
\textcolor{green}{\ding{51}}& \textcolor{red}{\ding{55}}  & \textcolor{red}{\ding{55}} &64.75\\
\textcolor{green}{\ding{51}}& \textcolor{green}{\ding{51}} &\textcolor{red}{\ding{55}}  &{65.85}\\
\textcolor{green}{\ding{51}}& \textcolor{green}{\ding{51}} &\textcolor{green}{\ding{51}}  &\textbf{66.35}\\
\hline
\end{tabular}
\end{table}

\begin{table}[t]
    \centering
      \caption{Cross-dataset performance comparison.}
      \label{tab:cross-data}
    \setlength{\tabcolsep}{1pt} 
    \renewcommand{\arraystretch}{0.7} 
    \begin{tabular}{lcc|cc|cc}
        \toprule
        \multirow{2}{*}{Methods} & \multicolumn{2}{c|}{Total-Text} & \multicolumn{2}{c|}{ICDAR13}& \multicolumn{2}{c}{BTS}\\
        \cmidrule(lr){2-3} \cmidrule(lr){4-5} \cmidrule(lr){6-7}
         & 4-shot & 16-shot & 4-shot & 16-shot   & 4-shot & 16-shot\\
        \midrule
        Visual & 55.4& 56.8&56.2 &57.6&51.9 &53.7 \\
        Visual+Prompt&56.7 &\textbf{58.1} &57.7 &\textbf{58.5}&53.6 &\textbf{55.2} \\
        \bottomrule
    \end{tabular}
\end{table}

\begin{figure}[t]
	\centering
\includegraphics[width=1.0\linewidth]{figure/visual4.pdf}
\vspace{-7mm}
         \caption{Heatmaps of TextSeg-trained model's activations on Total-Text/ICDAR13/BTS.}
         \label{fig:cross}
\end{figure}
 

\noindent\textbf{Qualitative comparison.} To further demonstrate the validity of our method, we compare the visualization results on the TextSeg~\cite{xu2021rethinking}, Total-Text~\cite{ch2017total}, and ICDAR13 FST~\cite{karatzas2013icdar} datasets under the 1-shot setting in Fig.~\ref{fig: case}.  Experimental results show that our method accurately segments text in complex scenarios, demonstrating greater robustness to background noise and achieving impressive performance.

\subsection{Component Effectiveness}
To investigate the effectiveness of each module in TSAL, we conduct extensive ablation experiments on TextSeg datasets,  including ablation studies on components, prompt templates and attribute settings. Furthermore, we validate TSAL's generalization capability through cross-dataset experiments comparing the two branches across three datasets: Total-Text, ICDAR13 FST, and BTS.

In Tab.~\ref{tab:abla}, we conduct ablation experiments on the model components. ``Visual'' refers to the visual-guided branch. ``Prompt'' refers to the adaptive prompt-guided branch. Using only the visual-guided branch corresponds to the zero-shot setting. Experiments show the adaptive prompt-guided branch achieves a 1.3\% improvement in FgIoU. This demonstrates that our designed prompts effectively capture the features of text and background. Applying the Adaptive Feature Alignment module~(AFA) achieves better performance, demonstrating AFA enables learnable prompts to better understand specific semantic information. 
The visualization results are shown in Fig.~\ref{fig: AFA}. We also investigate the effectiveness of enhancing foreground representations by incorporating background information.
For binary classification problems, fully utilizing background cues can improve the accuracy of foreground detection~\cite{liu2022learning, liu2020food, ye2015foreground}.
Experimental results show that introducing background information consistently improves performance, increasing FgIoU from 65.45 to 66.35 and AUROC from 93.14 to 95.11. This strategy effectively suppresses background noise and improves foreground localization.

 \begin{figure}[t]
	\centering
   \includegraphics[width=1.0\linewidth]{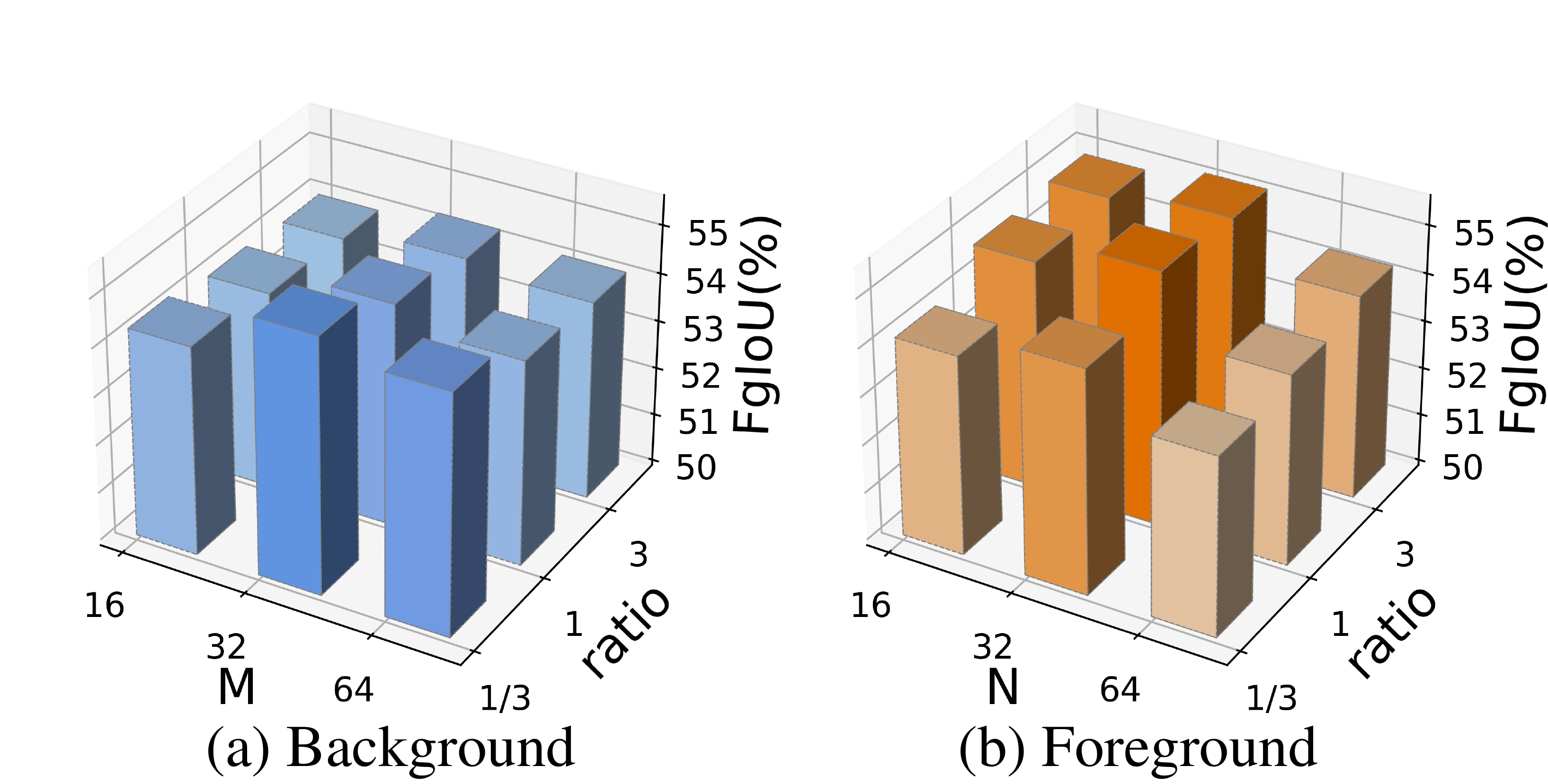}
        \caption{FgIoU of TextSeg~(1-shot) using different numbers of prompts and ratios of prototypes to learnable prompts.}
		\label{fig: back_num}
\vspace{-4mm}
\end{figure}

In Tab.~\ref{tab:fa} and Tab.~\ref{tab:ba}, we conduct experiments on the prompt templates of foreground and background prompts. The baseline uses the prompt template: ``a photo with XX''. For the foreground, the results demonstrate that each attribute designed for prompts effectively learns the corresponding information. Among them, the influence of color and position attributes is more significant, resulting in improvements of 2.21\% and 1.13\% in FgIoU, respectively. 
For the background, experiments show that dividing the background into distal and proximal scenes is effective for locating text regions and reducing noise. Generating fine-grained descriptions~(``Fine-grained'' in figures) for images to optimize the prompts also leads to improvements in model performance. Further details and visualizations regarding the attribute settings are provided in the supplementary material.

To verify that TSAL's prompt branch captures invariant text region representations beyond dataset-specific biases, we evaluate TSAL's cross-dataset robustness by training on TextSeg and testing on Total-Text, ICDAR13 FST, and BTS. Despite minor FgIoU drops from distribution shifts, TSAL achieves performance comparable to 16-shot adaptation as shown in Tab.~\ref{tab:cross-data}, demonstrating strong generalization. Ablation studies further confirm the effectiveness of our adaptive prompt branch, while qualitative visualizations~(Fig.~\ref{fig:cross}) reveal consistent prediction confidence across domains.

 \begin{figure}[t]
	\centering
   \includegraphics[width=1.0\linewidth]{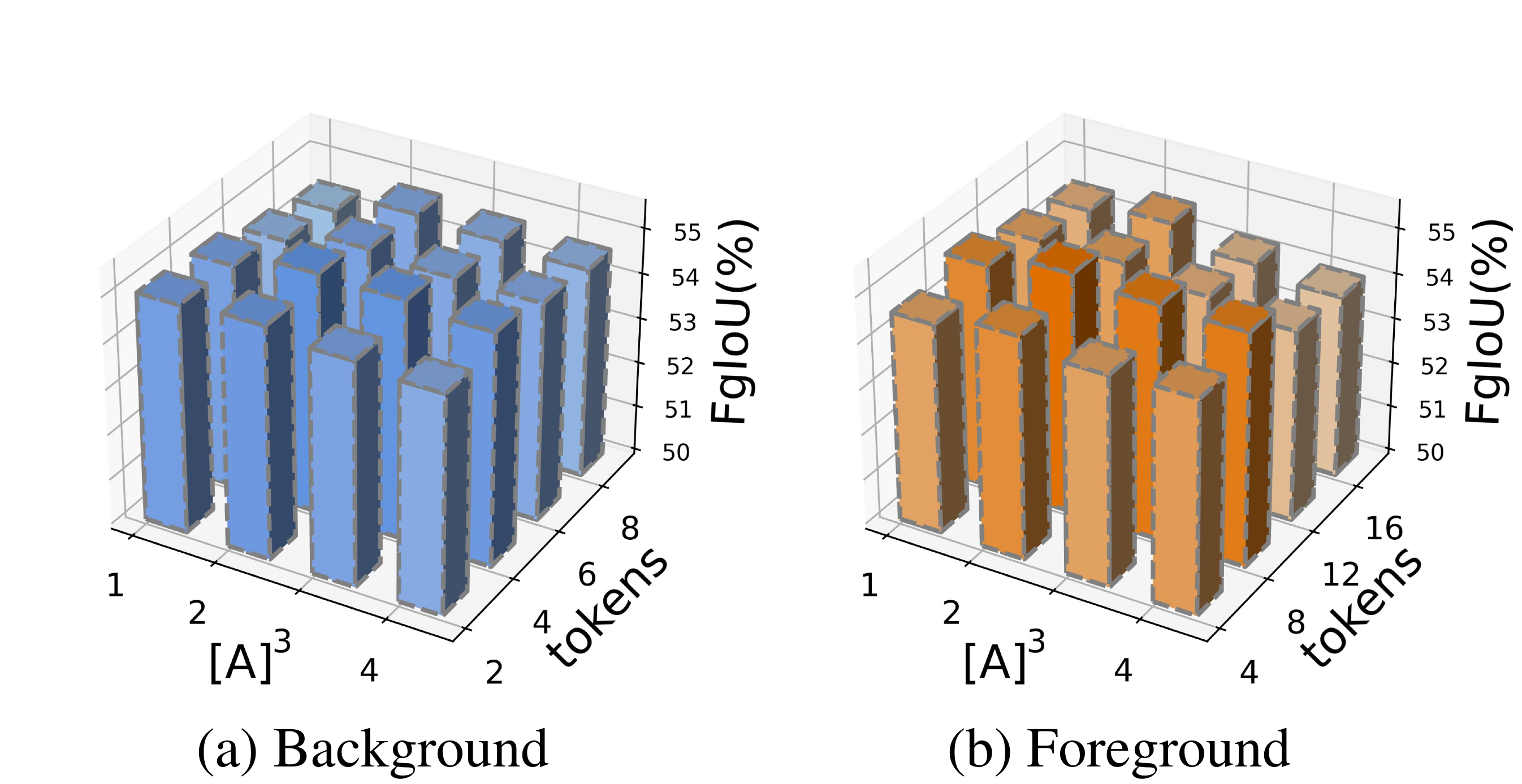}
        \caption{Results of FgIoU on TextSeg~(1-shot) setting using different numbers of $[A]$ and learnable tokens.}
		\label{fig: back_b}
\end{figure}

\begin{table}[t]
    \centering
    \caption{Ablation Study on the Effect of $\alpha$ and the Triplet Loss $L_{tri}$.}
    \label{tab:loss}
    \footnotesize
    \setlength{\tabcolsep}{3pt}
    \renewcommand{\arraystretch}{0.9}
    \begin{tabular}{lccccc|cc}  
        \toprule
        $\alpha$ & 0 & $10^{-1}$ & $10^{-2}$ & $10^{-3}$ & $10^{-4}$& $w/o$ $L_{tri}$&$w/$ $L_{tri}$ \\
        \midrule
        AUROC & 94.34 & 94.75 & \textbf{95.17} & 94.72 & 94.60& 93.34& \textbf{94.89}  \\
        \bottomrule
    \end{tabular}
    \vspace{-3mm}
\end{table}

\section{Discussion}
\label{discussion}
In this section, we conduct extensive ablation studies to analyze the impact of key hyperparameters on performance and robustness, and examine model design and efficiency aspects, including inference speed and memory consumption. These results provide comprehensive insights into the effectiveness and practicality of our design choices.

\subsection{Hyper-parameter Analysis}
\label{Sec:Hyper-parameter}
The number of prompts and learnable tokens plays a crucial role in model performance. In Fig.~\ref{fig: back_num}, we investigate the setting of the number of prompts. $M$ and $N$  represent the number of background and foreground prompts, respectively. ``ratio'' is the ratio of prompt prototypes to adaptive prompts. For the background, experiments show that $M=32$ \& $ratio=1/3$ produces better performance. For the foreground, $N=32$ \& $ratio=1$ produces better performance.
Too many learnable prompts hinder effective feature learning, while too few prompts reduces the robustness of the model.


\begin{table}[t]
    \centering
    \caption{Comparison of Model Complexity and Inference Efficiency.}
    \label{tab:key metrics}
     \vspace{-2mm}
    \footnotesize  
    \setlength{\tabcolsep}{3pt}  
    \renewcommand{\arraystretch}{0.9}  
    \begin{tabular}{lcccc}
        \toprule
        Methods & TFT & Hi-SAM-B & {TSAL} &SegGPT \\
        \midrule
        Params(M)/Mem(GB) & 7.61/8.2 & 16.73/14 & \textbf{0.12}/\textbf{2.8}& 371/-\\
        Latency(ms)/FPS & 80/12.5 & 31/32 & \textbf{12}/\textbf{85}&1900/- \\
        \bottomrule
    \end{tabular}
\end{table}
In Fig~\ref{fig: back_b}, we evaluate the effect of the number of $[A]$ and learnable tokens~(``tokens'' in figure) embedded in adaptive prompts. Results indicate that too many learnable tokens hinder prompt-prototype alignment, and adaptive prefixes $[A]$ fail to capture prior information from CLIP features. 
Too few learnable prefixes reduce prompt robustness.
Foreground prompts are more sensitive to the number of learnable tokens than background prompts, because they define multiple attributes to capture unique semantic and stylistic information, increasing the learning complexity.

\subsection{Ablation Study on the Loss Function}
We conduct ablation studies of $\alpha$ and $L_{tri}$ in the loss function, as shown in Tab.~\ref{tab:loss}. It can be observed that when $a = 0$ or larger, the performance degrades. This suggests that the distributions of visual features and prompt features should be aligned, but not excessively; otherwise, the diversity of the prompts is reduced, which in turn weakens the model's ability to perceive textual features. Using $L_{tri}$ yields better performance, as it enhances the model’s ability to discriminate between foreground and background.

\subsection{Computational Efficiency and Model Complexity}
We compare key metrics of different methods under the same settings using NVIDIA GeForce RTX 2080 Ti. The results are shown in Tab~\ref{tab:key metrics}. TSAL achieves impressive performance with minimal training cost.

Hi-SAM~\cite{ye2024hi} is a hierarchical text segmentation framework that leverages the Segment Anything Model~(SAM) to progressively extract text instances with structural awareness. Hi-SAM-B achieves 87.15\% fgIOU on TextSeg. While TSAL exhibits relatively modest performance, it offers significant advantages in terms of parameter efficiency, reduced training costs, and lower data dependency as shown in Tab~\ref{tab:key metrics}. Additionally, Hi-SAM builds upon SAM~(a segmentation-specialized model), whereas TSAL adapts CLIP, making direct comparisons unfair. 

While TSAL does not reach the performance of fully supervised models, it offers detector-free text localization with minimal labeling cost, making it suitable for low-resource scenarios, plug-in use, and annotation refinement.


\section{Conclusion}
We propose TSAL, the first few-shot scene text segmentation method that effectively incorporates CLIP's prior knowledge. TSAL consists of a visual-guided branch and an adaptive prompt-guided branch. The visual-guided branch extracts and stores foreground/background features for query matching, while the adaptive branch uses prompt templates and our Adaptive Feature Alignment~(AFA) module to capture diverse text attributes and complex backgrounds. TSAL is simple yet effective, achieving state-of-the-art results on public text segmentation benchmarks.


\bibliographystyle{ieeetr}
\bibliography{ref}

\end{document}